\documentclass[a4paper,fleqn]{cas-dc}

\usepackage[numbers]{natbib}
\usepackage{diagbox}

\usepackage{caption}
\usepackage[english]{babel}
\usepackage[utf8]{inputenc}
\usepackage{amsmath}
\usepackage{graphicx}
\usepackage[colorinlistoftodos]{todonotes}
\usepackage{hyperref}
\usepackage{color}
\usepackage{subfigure}
\usepackage{multirow}
\usepackage{geometry}
\usepackage{bbding}
\usepackage{amsmath}
\usepackage{graphicx}
\usepackage{cleveref}
\Crefname{figure}{Fig}{Figures}
\usepackage{float}

\def\tsc#1{\csdef{#1}{\textsc{\lowercase{#1}}\xspace}}
\tsc{WGM}
\tsc{QE}

\begin{document}
\captionsetup[figure]{labelfont={bf},labelformat={default},labelsep=period,name={Fig.}}
\let\WriteBookmarks\relax
\def\floatpagepagefraction{1}
\def\textpagefraction{.001}



\title [mode = title]{Explaining Model Overfitting in CNNs via GMM Clustering}  



%

\author[add1,add2]{Hui Dou}
\ead{huidou@smail.nju.edu.cn}
\author[add1,add4]{Xinyu Mu}
\author[add1,add4]{Mengjun Yi}
\author[add1,add2]{Feng Han}
\author[add3]{Jian Zhao}
\author[add1,add4]{Furao Shen\texorpdfstring{\corref{cor1}}{}}
\ead{frshen@nju.edu.cn}


\cortext[cor1]{Corresponding author.}

\address[add1]{State Key Laboratory for Novel Software Technology,
}
\address[add2]{Department of Computer Science and Technology, Nanjing University, Nanjing 210023, China,
}
\address[add3]{School of Electronic Science and Engineering, Nanjing University, Nanjing 210023, China,
}
\address[add4]{School of Artificial Intelligence, Nanjing University, Nanjing 210023, China,
}



\begin{abstract}
Convolutional Neural Networks (CNNs) have demonstrated remarkable prowess in the field of computer vision. 
However, their opaque decision-making processes pose significant challenges for practical applications.
In this study, we provide quantitative metrics for assessing CNN filters by clustering the feature maps corresponding to individual filters in the model via Gaussian Mixture Model (GMM). 
By analyzing the clustering results, we screen out some anomaly filters associated with outlier samples.
We further analyze the relationship between the anomaly filters and model overfitting, proposing three hypotheses.
This method is universally applicable across diverse CNN architectures without modifications, as evidenced by its successful application to models like AlexNet and LeNet-5. 
We present three meticulously designed experiments demonstrating our hypotheses from the perspectives of model behavior, dataset characteristics, and filter impacts. 
Through this work, we offer a novel perspective for evaluating the CNN performance and gain new insights into the operational behavior of model overfitting.
\end{abstract}




\maketitle







\section{Introduction}\label{section1}

In recent years, Convolutional Neural Networks (CNNs) have exhibited remarkable success across a multitude of applications, particularly in the domain of computer vision. 
Although CNNs exhibit impressive performance, they are frequently regarded as "black-box models", characterized by a lack of transparency and interpretability in their decision-making processes \cite{black-box}. 
This lack of transparency presents significant challenges, particularly in crucial domains like medical diagnosis, autonomous driving, and risk assessment, where a comprehensive understanding of model decisions is paramount for trust and reliability \cite{apply1, apply2, apply3}.

The interpretability of CNNs remains a pressing issue due to the absence of standardized methods. 
Various approaches have been proposed to elucidate the learning mechanisms of CNNs \cite{survey, survey2, survey3}. 
It is widely accepted that the convolutional filters in CNNs function as feature extractors \cite{multifaceted, feature}. 
While these extracted features are implicit to human understanding and need further processing, techniques such as visualization and spectral analysis can enhance our comprehension of them \cite{vis, spectral}. 
Typically, extracted features are multifaceted and intricately intertwined \cite{Graph}, with this entanglement becoming more pronounced as the model depth increases \cite{multifaceted}. Additionally, research suggests that filters in shallower layers tend to extract fundamental features such as shape, color, or texture, which are common across various classes. 
Conversely, filters in deeper layers tend to learn abstract concepts such as eyes, body parts and so on \cite{clustering}. 
These features are often more class-specific, thereby contributing to the CNN's classification capacity.

\begin{figure*}[htbp]

\centering
{
\includegraphics[width=0.8\textwidth]{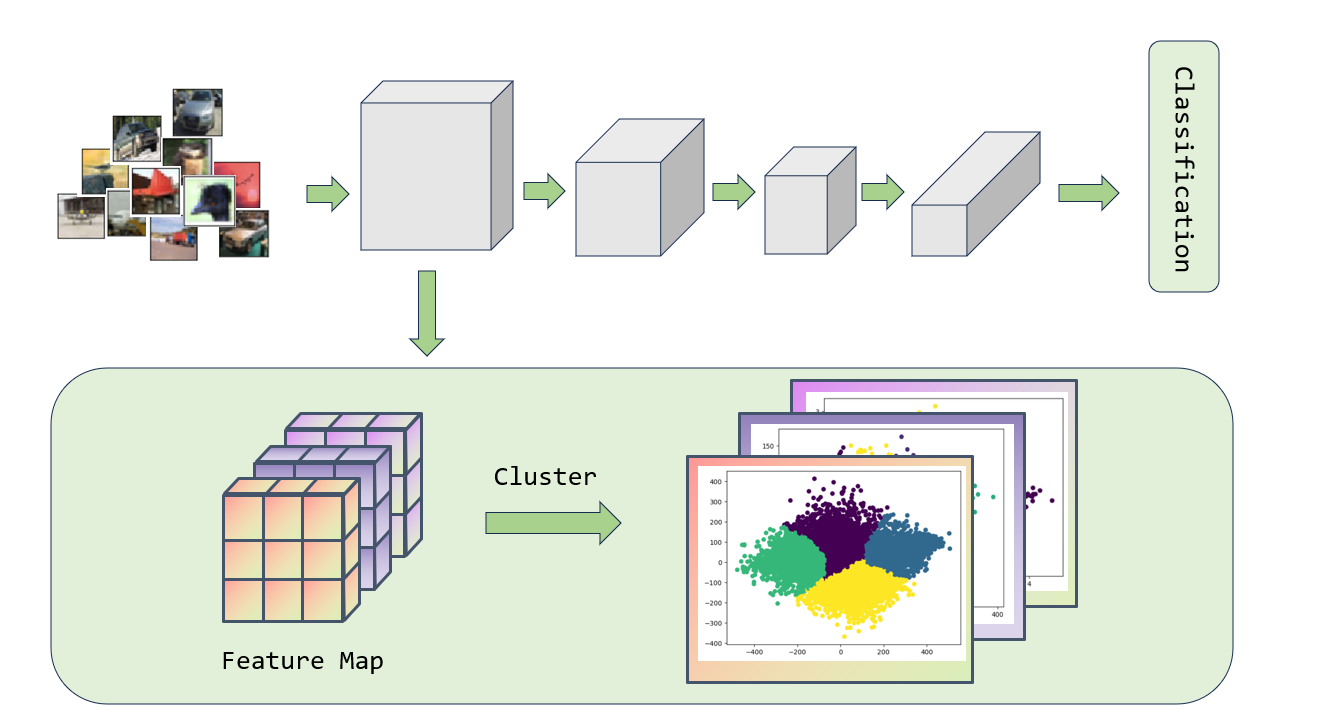}
}
\caption{
Given a pre-trained model, we cluster all the feature maps corresponding to the individual filter through the Gaussian Mixture Model.}
\label{fig:principle}
\end{figure*}

\begin{figure*}[htbp]

\centering
\subfigure[Normal Case 1]{
\includegraphics[width=0.7\textwidth]{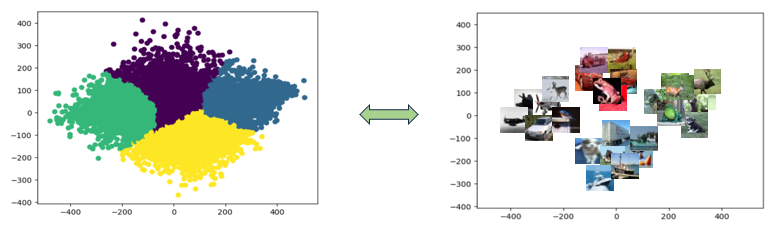}
\label{fig:normal1}
}
\quad
\subfigure[Normal Case 2]{
\includegraphics[width=0.7\textwidth]{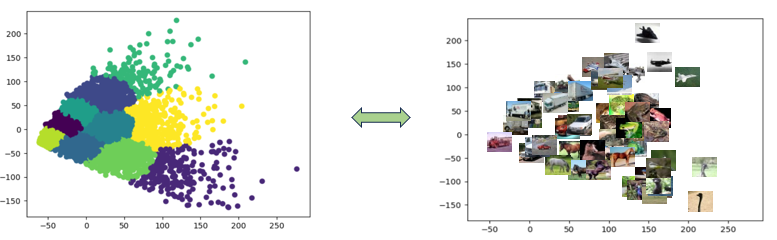}
\label{fig:normal2}
}
\quad
\subfigure[Rare Case]{
\includegraphics[width=0.7\textwidth]{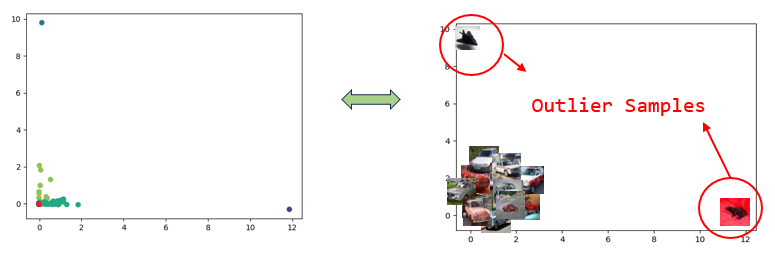}
\label{fig:rare1}
}
\quad
\caption{Visualization the clustering results. Each data point corresponds to one feature map generated by one input sample. 
There are normal cases where data points are evenly distributed and rare cases where outlier points occur.
The filter corresponding to the rare type of clustering result is the anomaly filter.}
\label{fig:outlier}
\end{figure*}

In this paper, we explain CNNs by exploring the relationship between the clustering results of feature maps and model overfitting.
Given a pre-trained model, we cluster all the feature maps corresponding to individual filters as shown in \Cref{fig:principle}. We further visualize the clustering results, where each data point corresponds to one feature map generated by one input image.
As illustrated in \Cref{fig:outlier}, in the normal cases, the feature maps exhibit effective clustering results, aligning with the conventional perspective that filters function as clustering functions.
In rare cases, small clusters or outliers points may appear in the clustering results.
For convenience, samples corresponding to outlier points and samples in small clusters are called \textit{outlier samples}.
We name the filter corresponding to this rare type of clustering result as \textit{anomaly filter}.
We find that the presence of the anomaly filter suggests potential overfitting of CNNs. 
Experiments are devised to validate this assumption, yielding the following hypotheses:
\begin{itemize}
\item[$\bullet$] \textbf{Anomaly filters increase in overfitting models.}
We observe models at different epochs during the training process and find a substantial increase in the number of anomaly filters when the model is overfitting.
\item[$\bullet$] \textbf{Outlier samples contribute to model overfitting.} 
We find that during the training process, the gradients of outlier samples are typically several times larger than those of normal ones.
This means that the model overlearns details in unusual samples, which may lead to overfitting of the model.
\item[$\bullet$] \textbf{Discarding anomaly filters enhances the generalization ability of the model.}
We compare the accuracy changes of the training datasets and validation datasets before and after pruning the anomaly filters in the overfitting models.
We find that pruning anomaly filters leads to a decrease in accuracy on the training datasets and an increase in accuracy on the validation datasets.
It means that after pruning the anomaly filters, the generalization ability of the model is improved.
\end{itemize}

In this paper, we primarily analyze the relationship between model overfitting and the anomaly filters in CNNs. The principal contributions of our method are outlined as follows:
\begin{itemize}
\item[$\bullet$] We provide fine-grained interpretation aligned with the CNN's nature.
We offer the explanations of CNNs at the filter level, in harmony with the inherent characteristics of CNNs. 
Leveraging unsupervised clustering, our method does not rely on predefined input features or prior human knowledge, thereby aligning interpretation more closely with the model's intrinsic nature.
\item[$\bullet$] We propose quantitative metrics to evaluate filters.
In order to assess CNNs at the filter level, we introduce an objective metric to detect anomaly filters related to model overfitting.
After clustering feature maps corresponding to individual filters, we employ an unsupervised metric to evaluate the quality of clustering results and further involve them in the selection of the anomaly filters.

\item[$\bullet$] We analyze the behavioral pattern of CNNs: 
The proposed method illuminates model behavior channel-by-channel and is adaptable to the CNN models without modification. 
By measuring clustering results, we identify a special pattern indicative of potential overfitting situations.
\end{itemize}

The organization of the paper is as follows: 
Section~\ref{section2} provides a brief survey of recent interpretation methods for CNNs from diverse perspectives. 
Section~\ref{section3} details the proposed method step-by-step, encompassing the selection of cluster methods, metrics, etc.
To elucidate the relationship between overfitting and the anomaly filter, we propose three meticulously designed experiments in Section~\ref{section4}, offering evidence from varied viewpoints. 
We incorporate three distinct models and datasets to corroborate the broad applicability of our method. 
Finally, we summarize our findings and offer a brief discussion in Section~\ref{section5}.

\section{Related Work}\label{section2}

Numerous methods have been proposed to improve the interpretability of CNNs, and certain patterns have been identified. 
For instance, researchers have observed that filters can function as feature extractors, learning intertwined and multifaceted features from images \cite{multifaceted, Graph}. 
Additionally, filters in shallower layers often learn generic features such as shape, color, and texture, whereas filters in deeper layers tend to extract abstract and discriminative concepts like eyes, body parts, etc \cite{clustering}. 
Here, we highlight some notable techniques:

\textbf{Network Visualization:} Visualization emerges as one of the most explicit and intuitive methods for unveiling hidden patterns in models. 
It encompasses two primary aspects: activation maximization and saliency map. 
Activation maximization generates inputs that maximize the activation for a specific neuron, thereby visualizing patterns learned by models. 
However, the images generated by activation maximization tend to be intricate and peculiar to human observers, necessitating techniques such as regularization to further optimize the results. 
\citet{multifaceted} introduced Multifaceted Feature Visualization and center-biased regularization to yield clearer and more comprehensive interpretability results. 
Saliency maps assign importance scores pixel-by-pixel. 
Class Activation Mapping (CAM) in \cite{CAM} generates a class activation map by multiplying the feature map of the last convolutional layer with relevant weights. 
However, it replaces the fully connected layer with global average pooling (GAP), thus requiring modification and retraining of the original models. 
Grad-CAM, introduced by \citet{Grad-CAM}, further incorporates gradients as the role of relevant weights. 
It requires no modification of models, only the backpropagation of gradients, thereby generating more flexible class activation maps. 
Layer-wise Relevance Propagation \cite{LWP} propagates the correlation of output back to quantify the contribution of each neuron/filter to the final prediction. 
It introduces some basic rules to obtain the decomposition of relevance in terms of messages sent to neurons of the previous layers. 
Neural Network Scanner (NNS) emulates the way fMRI works and scans given ANN models \cite{NNS}.
It is used to visualize neuron learning results for different components of neural networks in a unified way.

\textbf{Feature Attribution:} While saliency maps primarily assign attributions at the pixel level, other model-agnostic algorithms for feature attribution have been introduced. 
Local Interpretable Model-agnostic Explanations (LIME) \cite{LIME} is a technique for explaining classifier predictions in an interpretable and faithful manner. 
The main idea is to locally learn a model with better interpretability (such as a linear model or decision tree) around the prediction to serve as a substitute for the original model. 
Thus, LIME provides interpretation locally. 
SHapley Additive exPlanations (SHAP) \cite{SHAP} conducts feature attribution inspired by game theory. 
It introduces the Shapley value from game theory to calculate the contribution of relevant features, thus having a solid theoretical foundation.

\textbf{Semantic Information:} Algorithms aimed at extracting semantic information from models, which align better with human understanding, have been explored.
One approach involves incorporating human-defined knowledge. 
Testing with Concept Activation Vectors (TCAV) \cite{CAV} learns hyperplanes that separate samples with or without a certain concept, thereby quantifying the degree to which a user-defined concept influences a classification result. 
\citet{Graph} construct explanatory graphs for CNNs where each node corresponds to a specific concept. 
This method is based on the assumption of spatial relationships in CNNs. 
Another approach named Interpretable Convolutional Neural Network \cite{mask} involves modifying model architecture to mimic the way humans process information. 
It utilizes delicate masks and designed loss functions, enabling filters to learn single concepts instead of entangled combinations, making it more understandable to humans.

Among these methods, visualization is the most intuitive and explicit one. 
However, it is subjective due to the lack of common standards, often necessitating additional explanations from human observers. 
Feature attribution provides stronger theoretical foundations. 
Nevertheless, when applied to CNNs, it requires pre-defined artificial features such as superpixels. 
Moreover, solid theoretical foundations often result in algorithms with high complexity. 
Semantic information closely aligns with how we process visual input, but human-defined knowledge or artificial modifications may not align with the inherent nature of models, potentially introducing biases to interpretations. 
In other words, improvements in interpretability often come at the cost of performance or accuracy.

\section{Method}\label{section3}

\begin{figure*}[htbp]

\centering
{
\includegraphics[width=0.8\textwidth]{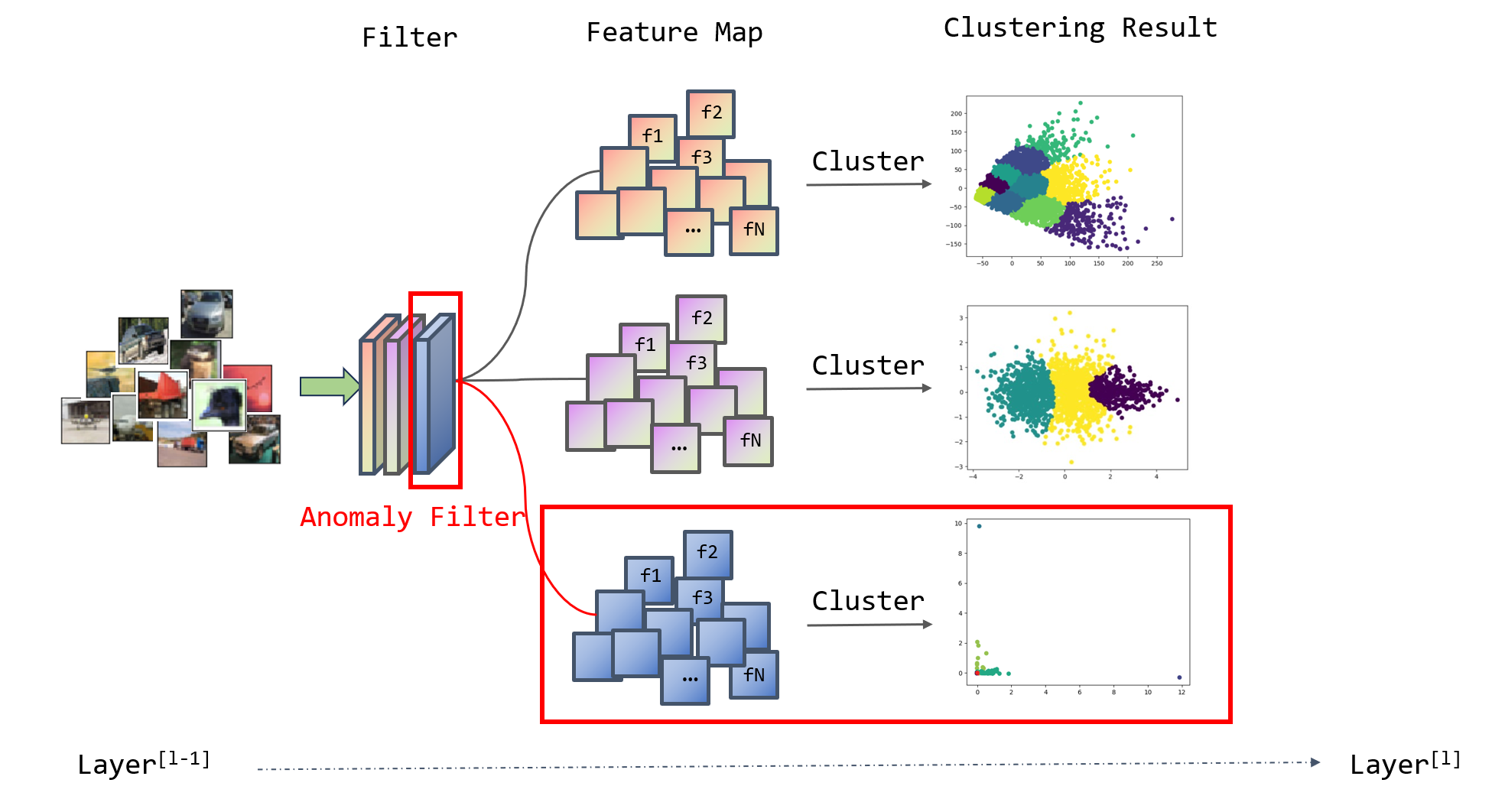}
}
\caption{We cluster the feature maps corresponding to the individual filter respectively, i.e., each data point in the clustering results corresponds to one feature map. Different colors represent different filters. Meanwhile, We categorize the clustering results into normal/rare cases. In this work, we mainly focus on the rare case scenario as marked in the red box.}
\label{fig:procedure}
\end{figure*}

In this section, we aim to evaluate and interpret CNNs at the filter level.
As shown in \Cref{fig:procedure}, we begin by clustering the feature maps corresponding to each filter using Gaussian Mixture Model (GMM).
The number of cluster classes $K$ is assigned meticulously and dynamically.
Subsequently, we establish criteria for identifying distinct patterns, referred to as the anomaly filters, which offer valuable insights into the CNN behavior. An anomaly filter possesses the following three defining attributes:
\begin{itemize}
\item[$\bullet$] Unbalanced class distribution.
In the normal filters, the distribution of data points within clusters tends to be relatively balanced.
However, in anomaly filters, the clustering results may show small clusters and outlier points.
The presence of these small clusters and outliers is the primary characteristic of the anomaly filter.
\item[$\bullet$] Abnormally high CH Index.
We utilize the Calinski-Harabasz Index (CH Index) \cite{CH} as a metric to evaluate the quality of clustering results. The anomaly filter demonstrates an abnormally high CH Index due to the presence of outlier points that are significantly distant from normal clusters.
\item[$\bullet$] Sufficiently large activation values.
This criterion ensures that the feature maps corresponding to the studied filters exhibit substantial activation values, indicating their relevance and warranting further investigation.
\end{itemize}

The detailed steps are listed as follows.

\subsection{Filter-based Clustering}
We begin by acquiring the feature maps $F^l$ of a certain layer $l$ consisting of $C$ channels and $F^l\in R^{Batch \times C \times W \times H} = [F^l_1,F^l_2,...,F^l_C]$.
To facilitate subsequent clustering, we reshape and reduce the dimension of the original $F^l$. 
Through concatenating, we obtain $S^l \in R^{Batch \times C \times (W*H)}$, on which Principal Component Analysis (PCA) is applied. 
Specifically, we reduce the dimension to 2 for further visualization and obtain 
$D^l \in R^{Batch \times C \times 2}$. 
Eventually, $D^l$ is acquired for the following clustering.

In this work, we employ a probability-based clustering algorithm called GMM. 
GMM assumes that data are drawn from a mixture of Gaussian distributions with different parameters $\pi_k, \mu_k, \Sigma_k$.
While $\mu_k$ and $\Sigma_k$ represent the mean and variance of the $k$th Gaussian distribution respectively, $\pi_k$ denotes the weight of the $k$th Gaussian distribution. 
Therefore, $ p(\boldsymbol{x})$ depicts the distribution where each data point $\boldsymbol{x}$ is sampled:

\begin{equation}
\label{eq:GMM}
 p\left(\boldsymbol{x} \right) = \sum_{k=1}^{K} \pi_k \mathcal{N}\left(\boldsymbol{x} | \mu_k, \Sigma_k\right).
\end{equation}

We apply GMM at the filter level.
For $D^l_c\in R^{Batch \times 2}$ in channel $c$, the classical Expectation-Maximization (EM) algorithm \cite{EM} is used for iterative optimization of relevant parameters.
For each data point $\boldsymbol{x}$ in the two-dimensional space, we initially calculate the posterior distribution of $\boldsymbol{x}$ belonging to a certain class:
\begin{equation}
\label{eq:posterior}
 p\left(\boldsymbol{z}=k|\boldsymbol{x}\right)=\frac{\pi_k\mathcal{N}\left(\boldsymbol{x}  | \mu_k, \Sigma_k\right)}{\sum_{k=1}^{K} \pi_k \mathcal{N}\left(\boldsymbol{x}  | \mu_k, \Sigma_k\right) },
\end{equation}
where $\boldsymbol{z}$ is a hidden variable denoting the class to which $\boldsymbol{x}$ belongs. 
Subsequently, the posterior distribution is used for reestimation of the marginal likelihood of the given data $D^l_c$ and parameter optimization:
\begin{equation}
\label{eq:likelihood}
\begin{aligned}
\quad \pi_k,\mu_k,\Sigma_k&=\mathop{\arg\max}\limits_{\pi_k,\mu_k,\Sigma_k} \sum_{\boldsymbol{x} } \ln\left(\sum_{k=1}^{K} \pi_k \mathcal{N}\left(\boldsymbol{x}  | \mu_k, \Sigma_k\right) \right).
\end{aligned}
\end{equation}

After several iterations, $\boldsymbol{x}$ is assigned to the class that maximizes the posterior distribution in Eq. \eqref{eq:posterior}. 
GMM takes advantage over traditional K-means, for GMM conducts soft clustering where the probability of data belonging to each class is calculated. 
Moreover, the nature of GMM allows it to fit clusters of all shapes and sizes, while K-means is limited to spherical clusters.
However, it should be noted that GMM still requires a preassigned class number $K$, which determines the total number of Gaussian distributions in Eq. \eqref{eq:GMM}. 
The selection of $K$ will be further discussed in Section~\ref{sec3.3}.

\subsection{Filter Evaluation}
Since the filter possesses the ability to cluster, it offers a novel perspective for exploring models by assessing the clustering results on feature maps. Naturally, filters that yield high-quality clustering results are deemed more significant to the model, as feature maps have likely captured common and useful information. Conversely, filters that produce poor-quality clustering results are considered to play a less impactful, or even detrimental, role in the model. Therefore, it is crucial to propose reasonable metrics for evaluating clustering.
Here we adopt the CH Index as our metrics:

\begin{equation}
\label{eq:CH index}
CH = \frac{SSB / \left(K - 1\right)}{SSW / \left(N - K\right)},
\end{equation}
where
\begin{equation}
\label{eq:SSW}
SSW = \sum_{i=1}^{K} \sum_{j=1}^{|C_i|} | \boldsymbol{x_{ij}} - \boldsymbol{m_i} |^2,
\end{equation}
and
\begin{equation}
\label{eq:SSB}
SSB = \sum_{i=1}^{K} |C_i| \cdot | \boldsymbol{m_i} - \boldsymbol{m} |^2.
\end{equation}

In Eq. \eqref{eq:CH index}, $SSB$ and $SSW$ stand for between-class and within-class divergence matrix respectively. 
While $SSB$ is calculated using the weighted Euclidean distance between each cluster center $\boldsymbol{m_i}$ and data center $\boldsymbol{m}$, $SSW$ takes the sum of Euclidean distance between each data point $\boldsymbol{x_i}$ and the corresponding cluster center $\boldsymbol{m_i}$. 
CH Index adopts an intuitive way of measuring the quality of clustering, wherein high-quality clusters should exhibit close resemblance within the same class while being distinguishable across different classes.

CH Index serves as a simple yet efficient metric for evaluating clustering in an unsupervised manner. 
In this work, we conduct filter-by-filter evaluations for the clustering results formed by $D^l\in R^{Batch \times C \times 2}$ and perform horizontal comparison within individual layers.

\subsection{Dynamic Assignment of the Number of Classes}\label{sec3.3}
In the aforementioned cluster methods, emphasis has been placed on the preassigned class number $K$. We have also illustrated that clusters of each filter indicate certain patterns it has learned, although the exact number remains elusive. 
To tackle this issue, we intend to dynamically assign the number of classes for each filter. 
All that is required is to determine an approximate range, from which we select $K$ as the final choice that yields the highest CH Index.
The reason is that the highest CH Index indicates the optimal fit of the selected $K$ to the behavior of filters, thus it can approximate the number of patterns learned.

\subsection{Anomaly Filter Selection}\label{sec3.4}
CH Index offers us quantitative filter evaluation. 
Through further visualization, a distinctive yet significant pattern in the clustering results is revealed. 
\Cref{fig:procedure} has shown some typical clustering results. 
In normal scenarios, clustering results exhibit the pattern characterized by evenly distributed data points.
However, a distinct pattern emerges where the majority of points cluster around the zero point (or a specific point), with a few outliers lying far away, thereby forming small clusters and outlier points.

There are three specific characteristics in the clustering results of an anomaly filter: 1. Unbalanced class distribution. 2. Abnormally high CH Index. 3. Sufficiently large activation values. 
As we visualize the clustering results of the anomaly filters in \Cref{fig:procedure}, we notice that a typical rare case scenario occurs when most of the data points cluster around zero point while only a few outliers are situated far away. While the majority of data form two or three clusters, outliers may form several individual clusters, thus contributing to characteristic 1. 
The characteristic 2 bears a close relation with the nature of CH Index itself. As observed in Eq. \eqref{eq:CH index}, an abnormally high CH Index may occur when $SSW$ is extremely small in magnitude and $SSB$ is significantly large. Eventually, to eliminate interference factors in the process of anomaly filter selection, we place a threshold on the magnitude of feature map activations, which efficiently excludes trivial situations where the filters are scarcely activated. The characteristic 3 aids in excluding redundant filters and shifting the focus towards filters that significantly impact model performance.
We consider the anomaly filter as a crucial indicator for analyzing the behavior of models, which will be explored further in the following section.


\section{Experiment}\label{section4}
In previous sections, we have introduced a novel method for CNN interpretation and evaluation on a channel-by-channel basis through clustering, and we have also emphasized the importance of the anomaly filter. 
Through meticulous analysis of the characteristics of the anomaly filters, we consider it as a potential indicator of overfitting, offering fresh insights into the CNN interpretation.
Here we devise three experiments to demonstrate the relationship between the anomaly filter and model overfitting. They are conducted across three CNN models and datasets.

\begin{table}[width=0.4\textwidth,pos=b]
\caption{Hyperparameters employed on experiments}\label{tab:para}
\begin{tabular*}{\tblwidth}{@{}LLLLL@{}}
\toprule
  \textbf{Experiment}& \textbf{${\lambda}$} & \textbf{${\alpha}$} &\textbf{${\beta}$}&\textbf{${\theta}$}\\   
\midrule
  \textbf{Experiment 1} &100& 0.2&1 &0.2\\
  \textbf{Experiment 2} &5 & -&- &-\\
  \textbf{Experiment 3} &50& 0.2 & 1.2 & 0.5\\

\bottomrule
\end{tabular*}
\end{table}

\textbf{Models:} To demonstrate the broad applicability of our method, we select three models of varying architectures. 
Our experimentation encompasses classical CNNs such as AlexNet \cite{Alexnet} and LeNet-5 \cite{LeNet}, alongside a three-layer \textit{simple CNN} with a filter configuration of 32-64-64. These models differ in depth, kernel size, channel number, etc., thereby validating the generalizability of our approach.

\textbf{Datasets:} In terms of datasets, we leverage the well-established CIFAR-10, CIFAR-100, and Fashion-MNIST, which are classical for image classification tasks. 
Larger datasets are excluded due to insufficient anomaly filters in the model to substantiate our findings. 
It should be noted that additional interpolation is necessary when applying these datasets to AlexNet. 
In this experiment, we employ the bicubic interpolation method.
It is a high-quality method to estimate pixel values through weighted averages of 16 neighboring pixels. 
This ensures conformity with AlexNet's input requirements, resulting in image sizes of 227x227x3.

\textbf{Hyperparameters:}
As mentioned in Section~\ref{sec3.4}, we delineate three criteria for identifying anomaly filters: 1. Unbalanced class distribution 2. Abnormally high CH Index 3. Sufficiently large activation values. We establish several hyperparameters for these criteria. For criterion 1, we designate clusters with no more than ${\lambda}$ points as small/anomaly clusters. If the number of anomaly clusters exceeds ${\alpha}$ times the total cluster number, the corresponding filter is deemed to exhibit an unbalanced class distribution. Regarding the abnormally high CH Index (criterion 2), we calculate the average CH Index for a given layer, designating filters with a CH Index exceeding ${\beta}$ times the average as meeting this criterion. For criterion 3, we compute the average activation of a given layer and set a threshold ${\theta}$ to exclude those inactive filters. That is to say, filters with activation values less than ${\theta}$ times average are off the table. Table \ref{tab:para} provides an overview of the hyperparameters employed across the three experiments.

\begin{figure*}[htbp]

\centering
\subfigure[The training curve]{
\includegraphics[width=0.4\textwidth]{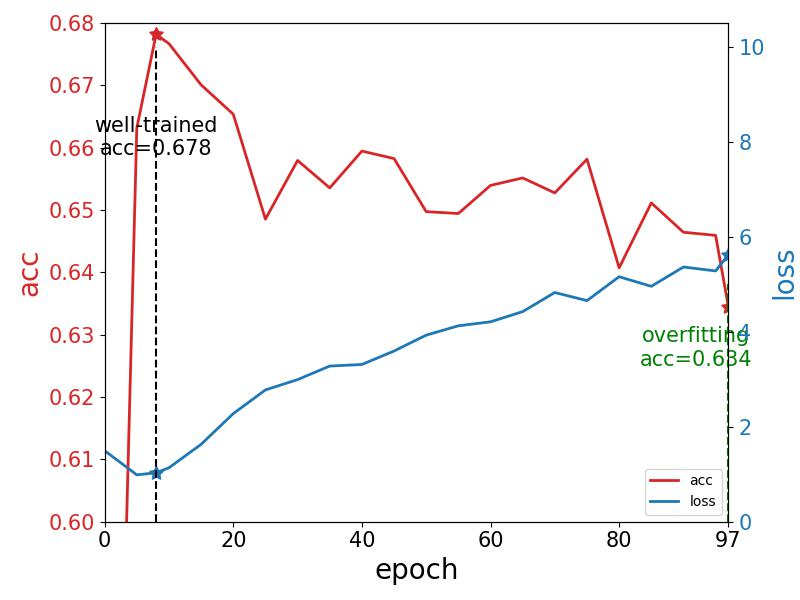}
\label{fig:curve1}
}
\quad
\subfigure[The number of anomaly filters]{
\includegraphics[width=0.44\textwidth]{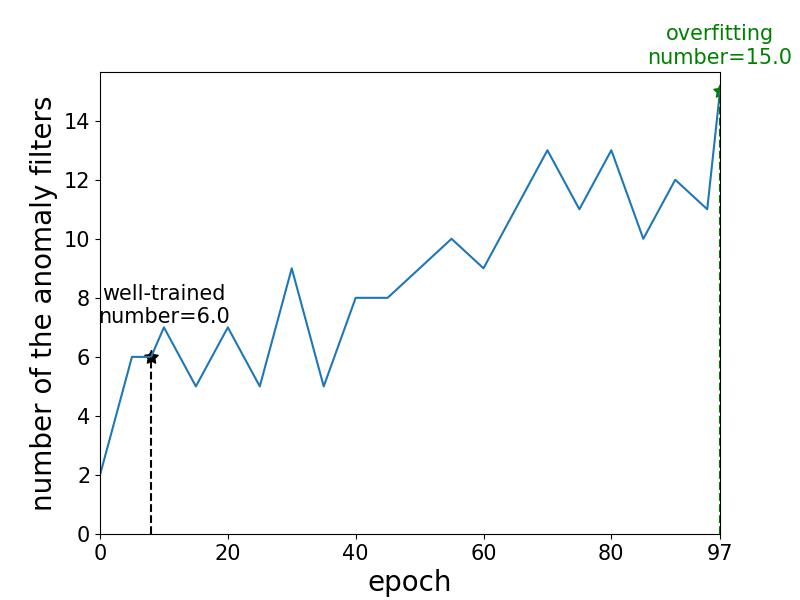}
\label{fig:curve2}
}
\quad

\caption{The training curve and the number of anomaly filters for a simple CNN in the dataset of CIFAR-10. Obvious overfitting occurs for there is a drop in the accuracy curve and an increase in the loss curve. The number of anomaly filters curve shows a similar trend to the loss curve, as when models become overfitting, the number of anomaly filters rises with fluctuations.}
\label{fig:curve}
\end{figure*}

\begin{table*}[width=0.8\textwidth,pos=tb]
\caption{The number of anomaly filters in models}\label{tab:ex1}
\begin{tabular*}{\tblwidth}{@{}LLLLLLL@{}}
\toprule
      \multirow{2}*{\textbf{\diagbox{Model}{Dataset}}}& \multicolumn{2}{c}{\textbf{CIFAR-10}}  & \multicolumn{2}{c}{\textbf{CIFAR-100}} & \multicolumn{2}{c}{\textbf{Fashion-MNIST}}\\
       &Well-trained & Overfitting &Well-trained & Overfitting &Well-trained &  Overfitting\\
\midrule    
      \textbf{Simple CNN} & 6  & \textbf{15} & 4 & \textbf{26} &1 & \textbf{5}\\
       \textbf{AlexNet} & 12 &\textbf{55} & 32 & \textbf{36} & 1&-\\
       \textbf{LeNet-5} &3 &\textbf{10} &11&\textbf{16} &1&\textbf{3} \\
\bottomrule     
\end{tabular*}
\end{table*}

\subsection{Anomaly filters increase in overfitting models}
To investigate the relationship between the anomaly filters and model overfitting, we focus on the variations in anomaly filter counts within the same model across different epochs.
We employ the CrossEntropy loss function and the Adam optimizer for training the models, with the epoch number set to 100.  
During the training process, we assess model overfitting based on the validation accuracy.
\Cref{fig:curve1} illustrates a typical training curve for a simple CNN on CIFAR-10. 
The accuracy on validation datasets initially ascends before descending, while the loss exhibits the inverse trend.
Additionally, we plot its anomaly filter curve in \Cref{fig:curve2}, depicting how the numbers of anomaly filters change in different epochs. We notice that the numbers of anomaly filters show a trend of fluctuation and rise, where the number of anomaly filters is relatively small in the well-trained epoch and large in the overfitting phase.

Additionally, we conduct a pairwise comparison by selecting a well-trained model and an overfitting model from each training process.
We designate well-trained models as those demonstrating generalization when the loss curve bottoms out and the accuracy curve peaks.
Conversely, for overfitting models, we select those from later epochs where there is a noticeable decline in validation accuracy.
As shown in \Cref{fig:curve1}, the well-trained model is highlighted in black and the overfitting model is highlighted in green.
Subsequently, we apply our algorithm to the well-trained models and the overfitting models respectively, and tabulate the numbers of anomaly filters. 
The result is listed in Table \ref{tab:ex1}. Notably, in this experimental setup, no apparent overfitting is observed for AlexNet on Fashion-MNIST. 
In other cases, we observe a higher incidence of anomaly filters in the overfitting models compared to their well-trained counterparts. Based on the aforementioned findings, we can confidently consider the anomaly filters as indicators of overfitting.

\subsection{Outlier samples contribute to model overfitting}
In this section, we investigate the relationship between the outlier samples and model overfitting. 
As mentioned in \cite{sharpness}, sharp minimizers of training functions precipitate overfitting. 
The overfitting models exhibit a propensity to adjust to minor fluctuations in training data, resulting in abnormally high gradients for certain parameters.
In our experiment, we quantify the impact of samples on model overfitting by calculating their gradient values.
As shown in Table \ref{tab:para}, we filter out samples that form clusters comprising no more than 5 data points as the outlier samples. 
This categorizes the data into outlier samples and normal samples.
Subsequently, we calculate the absolute gradient values on the well-trained models and the overfitting models respectively. 
Here we adopt the back-propagation algorithm, wherein the loss is propagated layer-by-layer in the form of gradients. 

We calculate the gradients of each layer and summarize their averages in Table \ref{tab:ex2}.
Firstly, we can observe that both the outlier samples and the normal samples exhibit higher gradients in the overfitting models compared with the well-trained models. 
Secondly, upon comparing the outlier samples with the normal samples, we observe that the outlier samples tend to have higher gradients in both the well-trained models and the overfitting models. For instance, in LeNet-5 trained on CIFAR-10, the gradients of outlier samples are over ten times higher than those of normal samples. This suggests that models develop more complex decision hyperplanes around outlier samples to accurately classify them, a characteristic symptom of overfitting.
Thus, we elucidate the relationship between overfitting and anomaly filters from the perspective of samples.

\begin{table*}[width=0.8\textwidth,pos=tb]
\caption{Mean gradients of the samples on different models}\label{tab:ex2}
\begin{tabular*}{\tblwidth}{@{}LLLLL@{}}
\toprule
      \multirow{2}*{\textbf{\diagbox{Model}{Dataset}}}& \multicolumn{2}{c}{\textbf{CIFAR-10}}  & \multicolumn{2}{c}{\textbf{CIFAR-100}}\\
        & Outlier Samples &Normal Samples & Outlier Samples &Normal Samples  \\
\midrule    
      \textbf{Simple CNN (well-trained)}& 0.063  & 0.022& 0.068& 0.019\\
      \textbf{Simple CNN (overfitting)} & 0.283  & 0.065& 0.384& 0.161\\
       \textbf{AlexNet (well-trained)} &0.005 &0.001 &0.010&0.002\\
       \textbf{AlexNet (overfitting)} & 0.013 &0.003 & 0.052&0.009\\
       \textbf{LeNet-5 (well-trained)} &0.455&0.048&0.149&0.033 \\
        \textbf{LeNet-5 (overfitting)} &2.758&0.221&0.261&0.161 \\
\bottomrule     
\end{tabular*}
\end{table*}

\begin{table*}[width=0.8\textwidth,pos=tb]
\caption{The proportion of filters that satisfy the decrease in training accuracy or increase in verification accuracy when anomaly filters are masked}\label{tab:ex3}
\begin{tabular*}{\tblwidth}{@{}LLLLL@{}}
\toprule
  \multirow{2}*{\textbf{\diagbox{Model}{Dataset}}}& \multicolumn{2}{c}{\textbf{CIFAR-10}}  & \multicolumn{2}{c}{\textbf{CIFAR-100}}\\   
  & Training ACC ↓ & Validation ACC ↑ & Training ACC ↓ &Validation ACC ↑  \\
\midrule
  \textbf{Simple CNN} & 3/4 &2/4& 10/10&8/10\\
  \textbf{AlexNet} & 5/7&5/7 & 13/19&12/19\\
  \textbf{LeNet-5} &3/3&1/3 & 1/1&0/1\\
\bottomrule
\end{tabular*}
\end{table*}

\begin{table*}[width=0.8\textwidth,pos=tb]
\caption{The accuracy fluctuation when a single filter is masked}\label{tab:ex3-3}
\begin{tabular*}{\tblwidth}{@{}LLLLL@{}}
\toprule
      \multirow{2}*{\textbf{\diagbox{Model}{Dataset}}}& \multicolumn{2}{c}{\textbf{CIFAR-10}}  & \multicolumn{2}{c}{\textbf{CIFAR-100}}\\
        & Anomaly Filter &Normal Filter & Anomaly Filter &Normal Filter  \\
\midrule    
      \textbf{Simple CNN} & \textbf{+0.03\%} & -1.19\% & \textbf{-0.31\%}& -1.12\%\\
       \textbf{AlexNet} & \textbf{+0.01\%} &-0.87\% & \textbf{+0.00\%}&-0.50\%\\
       \textbf{LeNet-5} &\textbf{-1.81\%}&-3.86\%&\textbf{-1.70\%}&-7.32\% \\
\bottomrule    
\end{tabular*}
\end{table*}

\begin{table}[width=0.4\textwidth,pos=tb]
\caption{The prediction results of outlier samples before/after anomaly filters being masked}\label{tab:ex4-2}
\begin{tabular*}{\tblwidth}{@{}LLL@{}}
\toprule
  \textbf{\diagbox{After}{Before}}& \textbf{True} & \textbf{False}\\   
\midrule
  \textbf{True} & 26 & 16\\
  \textbf{False} & 30 &107\\

\bottomrule
\end{tabular*}
\end{table}

\subsection{Discarding anomaly filters enhances the generalization of the model}

In this section, we conducted a pruning experiment to assess the impact of anomaly filters on the model's generalization ability. 
The anomaly filter was masked by setting its values to zero. 
If an anomaly occurs on the ReLU or maxpool layer, we simply mask the corresponding upper convolutional filter. 
In this way, we conduct masking at the filter level. 
Accuracy on the validation dataset is calculated to evaluate the impact of the masked filter on the entire model.
It should be noted that stricter rules have been applied in this experiment to screen out the anomaly filter, as masking too many filters will inevitably lead to a decline in accuracy. Consequently, there is no anomaly filter on Fashion-MNIST.

We first conduct masking on a single filter (single-filter masking) and summarize the results in Table \ref{tab:ex3}.
Table \ref{tab:ex3} shows the percentage of rise in validation accuracy and that of drop in training accuracy due to single-filter masking respectively. We can observe that the majority of single-filter maskings directly increase the validation accuracy while decreasing the training accuracy, indicating improved generalization ability of the masked models.
Better results are possible when we apply stricter rules when screening out the anomaly filter. 
Additional experiment is carried out to further verify the effectiveness of single-filter masking of the anomaly filters. 
We carry out random single-filter masking and summarize the results in Table \ref{tab:ex3-3}.
Notably, single-filter maskings of the anomaly filters result in fewer accuracy drops compared to random maskings in all scenarios.
Thus, the experiment demonstrates that removing anomaly filters helps enhance the model's generalization ability.

In addition, we take a simple CNN on CIFAR-100 as an example for detailed analysis. 
We mask anomaly filters on the model. 
As depicted in Table \ref{tab:ex4-2}, upon applying the masked models to the outlier samples, a notable alteration in the classification results is observed. 
We can see that most outlier samples are misclassified samples.
Among 179 outlier samples, 107 outlier samples are always misclassified before and after the masking operation. 
Only 26 correctly classified outlier samples maintain their respective classifications. 
The outlier samples are sensitive to changes in the model fitting curve.
30 out of 179 outlier samples shift from being correctly classified to being misclassified and 16 outlier samples are reclassified from misclassification to correct classification.
It indicates that the drop in accuracy is primarily caused by misclassification of the outlier samples, ultimately contributing to improved generalization ability. 
In this way, we clarify the relationship between the anomaly filters and the generalization ability of the model by conducting pruning.



\section{Conclusion and Future work}\label{section5}
In this paper, we present a novel method to investigate the relationship between CNN filters and model overfitting, It is compatible with the nature of CNNs by adopting the idea of GMM clustering. By clustering the feature maps corresponding to individual filters, we identify the anomaly filters that exhibit a close correlation with model overfitting. We propose three hypotheses:
1. Anomaly filters increase in overfitting models.
2. Outlier samples contribute to model overfitting.
3. Removal of anomaly filters enhances the generalization ability of the model.
To validate these hypotheses, we design three experiments. By incorporating CNN models of different architectures, the method is proven to have broad applicability. The architectures utilized in this study are relatively primitive. Future investigations will explore the application of our method to more sophisticated CNN models such as ResNet, aiming to uncover additional underlying patterns. The inclusion of residual modules in ResNet introduces complexities that pose significant challenges to further analysis. Additionally, we will explore novel metrics for evaluating clustering results, which would better incorporate semantic information.

\section{Acknowledgment}
This work was supported in part by the STI 2030-Major Projects of China under Grant 2021ZD0201300, and by the National Science Foundation of China under Grant 62276127.









\bibliographystyle{elsarticle-num-names} 
\bibliography{paper1}

\begin{thebibliography}{26}
\expandafter\ifx\csname natexlab\endcsname\relax\def\natexlab#1{#1}\fi
\providecommand{\url}[1]{\texttt{#1}}
\providecommand{\href}[2]{#2}
\providecommand{\path}[1]{#1}
\providecommand{\DOIprefix}{doi:}
\providecommand{\ArXivprefix}{arXiv:}
\providecommand{\URLprefix}{URL: }
\providecommand{\Pubmedprefix}{pmid:}
\providecommand{\doi}[1]{\href{http://dx.doi.org/#1}{\path{#1}}}
\providecommand{\Pubmed}[1]{\href{pmid:#1}{\path{#1}}}
\providecommand{\bibinfo}[2]{#2}
\ifx\xfnm\relax \def\xfnm[#1]{\unskip,\space#1}\fi
\bibitem[{Rudin(2019)}]{black-box}
\bibinfo{author}{C.~Rudin},
\newblock \bibinfo{title}{Stop explaining black box machine learning models for high stakes decisions and use interpretable models instead},
\newblock \bibinfo{journal}{Nat Mach Intell} \bibinfo{volume}{1} (\bibinfo{year}{2019}) \bibinfo{pages}{206--215}. \URLprefix \url{https://www.nature.com/articles/s42256-019-0048-x}. \DOIprefix\doi{10.1038/s42256-019-0048-x}.
\bibitem[{Rudin et~al.(2021)Rudin, Chen, Chen, Huang, Semenova, and Zhong}]{apply1}
\bibinfo{author}{C.~Rudin}, \bibinfo{author}{C.~Chen}, \bibinfo{author}{Z.~Chen}, \bibinfo{author}{H.~Huang}, \bibinfo{author}{L.~Semenova}, \bibinfo{author}{C.~Zhong}, \bibinfo{title}{Interpretable {Machine} {Learning}: {Fundamental} {Principles} and 10 {Grand} {Challenges}}, \bibinfo{year}{2021}. \URLprefix \url{http://arxiv.org/abs/2103.11251}, \bibinfo{note}{arXiv:2103.11251 [cs, stat]}.
\bibitem[{Zhu et~al.(2022)Zhu, Ma, Yuan, and Zhu}]{apply2}
\bibinfo{author}{Y.~Zhu}, \bibinfo{author}{J.~Ma}, \bibinfo{author}{C.~Yuan}, \bibinfo{author}{X.~Zhu},
\newblock \bibinfo{title}{Interpretable learning based {Dynamic} {Graph} {Convolutional} {Networks} for {Alzheimer}’s {Disease} analysis},
\newblock \bibinfo{journal}{Information Fusion} \bibinfo{volume}{77} (\bibinfo{year}{2022}) \bibinfo{pages}{53--61}. \URLprefix \url{https://linkinghub.elsevier.com/retrieve/pii/S1566253521001548}. \DOIprefix\doi{10.1016/j.inffus.2021.07.013}.
\bibitem[{You et~al.(2020)You, Leskovec, He, and Xie}]{apply3}
\bibinfo{author}{J.~You}, \bibinfo{author}{J.~Leskovec}, \bibinfo{author}{K.~He}, \bibinfo{author}{S.~Xie},
\newblock \bibinfo{title}{Graph structure of neural networks},
\newblock in: \bibinfo{editor}{H.~D. III}, \bibinfo{editor}{A.~Singh} (Eds.), \bibinfo{booktitle}{Proceedings of the 37th International Conference on Machine Learning}, volume \bibinfo{volume}{119} of \textit{\bibinfo{series}{Proceedings of Machine Learning Research}}, \bibinfo{publisher}{PMLR}, \bibinfo{year}{2020}, pp. \bibinfo{pages}{10881--10891}. \URLprefix \url{https://proceedings.mlr.press/v119/you20b.html}.
\bibitem[{Zhang et~al.(2021)Zhang, Tiňo, Leonardis, and Tang}]{survey}
\bibinfo{author}{Y.~Zhang}, \bibinfo{author}{P.~Tiňo}, \bibinfo{author}{A.~Leonardis}, \bibinfo{author}{K.~Tang},
\newblock \bibinfo{title}{A {Survey} on {Neural} {Network} {Interpretability}},
\newblock \bibinfo{journal}{IEEE Trans. Emerg. Top. Comput. Intell.} \bibinfo{volume}{5} (\bibinfo{year}{2021}) \bibinfo{pages}{726--742}. \URLprefix \url{http://arxiv.org/abs/2012.14261}. \DOIprefix\doi{10.1109/TETCI.2021.3100641}, \bibinfo{note}{arXiv:2012.14261 [cs]}.
\bibitem[{Gilpin et~al.(2019)Gilpin, Bau, Yuan, Bajwa, Specter, and Kagal}]{survey2}
\bibinfo{author}{L.~H. Gilpin}, \bibinfo{author}{D.~Bau}, \bibinfo{author}{B.~Z. Yuan}, \bibinfo{author}{A.~Bajwa}, \bibinfo{author}{M.~Specter}, \bibinfo{author}{L.~Kagal}, \bibinfo{title}{Explaining {Explanations}: {An} {Overview} of {Interpretability} of {Machine} {Learning}}, \bibinfo{year}{2019}. \URLprefix \url{http://arxiv.org/abs/1806.00069}, \bibinfo{note}{arXiv:1806.00069 [cs, stat]}.
\bibitem[{Montavon et~al.(2018)Montavon, Samek, and Müller}]{survey3}
\bibinfo{author}{G.~Montavon}, \bibinfo{author}{W.~Samek}, \bibinfo{author}{K.-R. Müller},
\newblock \bibinfo{title}{Methods for interpreting and understanding deep neural networks},
\newblock \bibinfo{journal}{Digital Signal Processing} \bibinfo{volume}{73} (\bibinfo{year}{2018}) \bibinfo{pages}{1--15}. \URLprefix \url{https://linkinghub.elsevier.com/retrieve/pii/S1051200417302385}. \DOIprefix\doi{10.1016/j.dsp.2017.10.011}.
\bibitem[{Nguyen et~al.(2016)Nguyen, Yosinski, and Clune}]{multifaceted}
\bibinfo{author}{A.~Nguyen}, \bibinfo{author}{J.~Yosinski}, \bibinfo{author}{J.~Clune}, \bibinfo{title}{Multifaceted {Feature} {Visualization}: {Uncovering} the {Different} {Types} of {Features} {Learned} {By} {Each} {Neuron} in {Deep} {Neural} {Networks}}, \bibinfo{year}{2016}. \URLprefix \url{http://arxiv.org/abs/1602.03616}, \bibinfo{note}{arXiv:1602.03616 [cs]}.
\bibitem[{Athiwaratkun and Kang(2015)}]{feature}
\bibinfo{author}{B.~Athiwaratkun}, \bibinfo{author}{K.~Kang}, \bibinfo{title}{Feature {Representation} in {Convolutional} {Neural} {Networks}}, \bibinfo{year}{2015}. \URLprefix \url{http://arxiv.org/abs/1507.02313}. \DOIprefix\doi{10.48550/arXiv.1507.02313}, \bibinfo{note}{arXiv:1507.02313 [cs]}.
\bibitem[{Dosovitskiy and Brox(2016)}]{vis}
\bibinfo{author}{A.~Dosovitskiy}, \bibinfo{author}{T.~Brox},
\newblock \bibinfo{title}{Inverting {Visual} {Representations} with {Convolutional} {Networks}},
\newblock in: \bibinfo{booktitle}{2016 {IEEE} {Conference} on {Computer} {Vision} and {Pattern} {Recognition} ({CVPR})}, \bibinfo{publisher}{IEEE}, \bibinfo{address}{Las Vegas, NV, USA}, \bibinfo{year}{2016}, pp. \bibinfo{pages}{4829--4837}. \URLprefix \url{http://ieeexplore.ieee.org/document/7780891/}. \DOIprefix\doi{10.1109/CVPR.2016.522}.
\bibitem[{Zhang et~al.(2020)Zhang, Xu, Yang, Chen, Zhou, Liu, Li, Lin, and Ying}]{spectral}
\bibinfo{author}{X.~Zhang}, \bibinfo{author}{J.~Xu}, \bibinfo{author}{J.~Yang}, \bibinfo{author}{L.~Chen}, \bibinfo{author}{H.~Zhou}, \bibinfo{author}{X.~Liu}, \bibinfo{author}{H.~Li}, \bibinfo{author}{T.~Lin}, \bibinfo{author}{Y.~Ying},
\newblock \bibinfo{title}{Understanding the learning mechanism of convolutional neural networks in spectral analysis},
\newblock \bibinfo{journal}{Analytica Chimica Acta} \bibinfo{volume}{1119} (\bibinfo{year}{2020}) \bibinfo{pages}{41--51}. \URLprefix \url{https://linkinghub.elsevier.com/retrieve/pii/S0003267020303767}. \DOIprefix\doi{10.1016/j.aca.2020.03.055}.
\bibitem[{Zhang et~al.(2021)Zhang, Wang, Cao, Wu, Shi, and Zhu}]{Graph}
\bibinfo{author}{Q.~Zhang}, \bibinfo{author}{X.~Wang}, \bibinfo{author}{R.~Cao}, \bibinfo{author}{Y.~N. Wu}, \bibinfo{author}{F.~Shi}, \bibinfo{author}{S.-C. Zhu},
\newblock \bibinfo{title}{Extraction of an {Explanatory} {Graph} to {Interpret} a {CNN}},
\newblock \bibinfo{journal}{IEEE Trans. Pattern Anal. Mach. Intell.} \bibinfo{volume}{43} (\bibinfo{year}{2021}) \bibinfo{pages}{3863--3877}. \URLprefix \url{https://ieeexplore.ieee.org/document/9086075/}. \DOIprefix\doi{10.1109/TPAMI.2020.2992207}.
\bibitem[{Girish et~al.(2019)Girish, Singh, and Ralescu}]{clustering}
\bibinfo{author}{D.~Girish}, \bibinfo{author}{V.~Singh}, \bibinfo{author}{A.~L. Ralescu},
\newblock \bibinfo{title}{Unsupervised clustering based understanding of cnn.},
\newblock in: \bibinfo{booktitle}{CVPR Workshops}, \bibinfo{year}{2019}, pp. \bibinfo{pages}{9--11}.
\bibitem[{Zhou et~al.(2016)Zhou, Khosla, Lapedriza, Oliva, and Torralba}]{CAM}
\bibinfo{author}{B.~Zhou}, \bibinfo{author}{A.~Khosla}, \bibinfo{author}{A.~Lapedriza}, \bibinfo{author}{A.~Oliva}, \bibinfo{author}{A.~Torralba},
\newblock \bibinfo{title}{Learning {Deep} {Features} for {Discriminative} {Localization}},
\newblock in: \bibinfo{booktitle}{2016 {IEEE} {Conference} on {Computer} {Vision} and {Pattern} {Recognition} ({CVPR})}, \bibinfo{publisher}{IEEE}, \bibinfo{address}{Las Vegas, NV, USA}, \bibinfo{year}{2016}, pp. \bibinfo{pages}{2921--2929}. \URLprefix \url{http://ieeexplore.ieee.org/document/7780688/}. \DOIprefix\doi{10.1109/CVPR.2016.319}.
\bibitem[{Selvaraju et~al.(2020)Selvaraju, Cogswell, Das, Vedantam, Parikh, and Batra}]{Grad-CAM}
\bibinfo{author}{R.~R. Selvaraju}, \bibinfo{author}{M.~Cogswell}, \bibinfo{author}{A.~Das}, \bibinfo{author}{R.~Vedantam}, \bibinfo{author}{D.~Parikh}, \bibinfo{author}{D.~Batra},
\newblock \bibinfo{title}{Grad-{CAM}: {Visual} {Explanations} from {Deep} {Networks} via {Gradient}-based {Localization}},
\newblock \bibinfo{journal}{Int J Comput Vis} \bibinfo{volume}{128} (\bibinfo{year}{2020}) \bibinfo{pages}{336--359}. \URLprefix \url{http://arxiv.org/abs/1610.02391}. \DOIprefix\doi{10.1007/s11263-019-01228-7}, \bibinfo{note}{arXiv:1610.02391 [cs]}.
\bibitem[{Bach et~al.(2015)Bach, Binder, Montavon, Klauschen, Müller, and Samek}]{LWP}
\bibinfo{author}{S.~Bach}, \bibinfo{author}{A.~Binder}, \bibinfo{author}{G.~Montavon}, \bibinfo{author}{F.~Klauschen}, \bibinfo{author}{K.-R. Müller}, \bibinfo{author}{W.~Samek},
\newblock \bibinfo{title}{On {Pixel}-{Wise} {Explanations} for {Non}-{Linear} {Classifier} {Decisions} by {Layer}-{Wise} {Relevance} {Propagation}},
\newblock \bibinfo{journal}{PLoS ONE} \bibinfo{volume}{10} (\bibinfo{year}{2015}) \bibinfo{pages}{e0130140}. \URLprefix \url{https://dx.plos.org/10.1371/journal.pone.0130140}. \DOIprefix\doi{10.1371/journal.pone.0130140}.
\bibitem[{Dou et~al.(2023)Dou, Shen, Zhao, and Mu}]{NNS}
\bibinfo{author}{H.~Dou}, \bibinfo{author}{F.~Shen}, \bibinfo{author}{J.~Zhao}, \bibinfo{author}{X.~Mu},
\newblock \bibinfo{title}{Understanding neural network through neuron level visualization},
\newblock \bibinfo{journal}{Neural Networks} \bibinfo{volume}{168} (\bibinfo{year}{2023}) \bibinfo{pages}{484--495}. \URLprefix \url{https://linkinghub.elsevier.com/retrieve/pii/S0893608023005269}. \DOIprefix\doi{10.1016/j.neunet.2023.09.030}.
\bibitem[{Ribeiro et~al.(2016)Ribeiro, Singh, and Guestrin}]{LIME}
\bibinfo{author}{M.~T. Ribeiro}, \bibinfo{author}{S.~Singh}, \bibinfo{author}{C.~Guestrin}, \bibinfo{title}{"{Why} {Should} {I} {Trust} {You}?": {Explaining} the {Predictions} of {Any} {Classifier}}, \bibinfo{year}{2016}. \URLprefix \url{http://arxiv.org/abs/1602.04938}, \bibinfo{note}{arXiv:1602.04938 [cs, stat]}.
\bibitem[{Lundberg and Lee(2017)}]{SHAP}
\bibinfo{author}{S.~M. Lundberg}, \bibinfo{author}{S.-I. Lee},
\newblock \bibinfo{title}{A unified approach to interpreting model predictions},
\newblock \bibinfo{journal}{Advances in neural information processing systems} \bibinfo{volume}{30} (\bibinfo{year}{2017}).
\bibitem[{Kim et~al.(2018)Kim, Wattenberg, Gilmer, Cai, Wexler, Viegas, and Sayres}]{CAV}
\bibinfo{author}{B.~Kim}, \bibinfo{author}{M.~Wattenberg}, \bibinfo{author}{J.~Gilmer}, \bibinfo{author}{C.~Cai}, \bibinfo{author}{J.~Wexler}, \bibinfo{author}{F.~Viegas}, \bibinfo{author}{R.~Sayres}, \bibinfo{title}{Interpretability {Beyond} {Feature} {Attribution}: {Quantitative} {Testing} with {Concept} {Activation} {Vectors} ({TCAV})}, \bibinfo{year}{2018}. \URLprefix \url{http://arxiv.org/abs/1711.11279}, \bibinfo{note}{arXiv:1711.11279 [stat]}.
\bibitem[{Zhang et~al.(2018)Zhang, Wu, and Zhu}]{mask}
\bibinfo{author}{Q.~Zhang}, \bibinfo{author}{Y.~N. Wu}, \bibinfo{author}{S.-C. Zhu}, \bibinfo{title}{Interpretable {Convolutional} {Neural} {Networks}}, \bibinfo{year}{2018}. \URLprefix \url{http://arxiv.org/abs/1710.00935}, \bibinfo{note}{arXiv:1710.00935 [cs]}.
\bibitem[{Calinski and Harabasz(1974)}]{CH}
\bibinfo{author}{T.~Calinski}, \bibinfo{author}{J.~Harabasz},
\newblock \bibinfo{title}{A dendrite method for cluster analysis},
\newblock \bibinfo{journal}{Comm. in Stats. - Theory \& Methods} \bibinfo{volume}{3} (\bibinfo{year}{1974}) \bibinfo{pages}{1--27}. \URLprefix \url{http://www.tandfonline.com/doi/abs/10.1080/03610927408827101}. \DOIprefix\doi{10.1080/03610927408827101}.
\bibitem[{Dempster et~al.(1977)Dempster, Laird, and Rubin}]{EM}
\bibinfo{author}{A.~P. Dempster}, \bibinfo{author}{N.~M. Laird}, \bibinfo{author}{D.~B. Rubin},
\newblock \bibinfo{title}{Maximum likelihood from incomplete data via the em algorithm},
\newblock \bibinfo{journal}{Journal of the royal statistical society: series B (methodological)} \bibinfo{volume}{39} (\bibinfo{year}{1977}) \bibinfo{pages}{1--22}.
\bibitem[{Krizhevsky et~al.(2017)Krizhevsky, Sutskever, and Hinton}]{Alexnet}
\bibinfo{author}{A.~Krizhevsky}, \bibinfo{author}{I.~Sutskever}, \bibinfo{author}{G.~E. Hinton},
\newblock \bibinfo{title}{{ImageNet} classification with deep convolutional neural networks},
\newblock \bibinfo{journal}{Commun. ACM} \bibinfo{volume}{60} (\bibinfo{year}{2017}) \bibinfo{pages}{84--90}. \URLprefix \url{https://dl.acm.org/doi/10.1145/3065386}. \DOIprefix\doi{10.1145/3065386}.
\bibitem[{Lecun et~al.(1998)Lecun, Bottou, Bengio, and Haffner}]{LeNet}
\bibinfo{author}{Y.~Lecun}, \bibinfo{author}{L.~Bottou}, \bibinfo{author}{Y.~Bengio}, \bibinfo{author}{P.~Haffner},
\newblock \bibinfo{title}{Gradient-based learning applied to document recognition},
\newblock \bibinfo{journal}{Proc. IEEE} \bibinfo{volume}{86} (\bibinfo{year}{1998}) \bibinfo{pages}{2278--2324}. \URLprefix \url{http://ieeexplore.ieee.org/document/726791/}. \DOIprefix\doi{10.1109/5.726791}.
\bibitem[{Keskar et~al.(2016)Keskar, Mudigere, Nocedal, Smelyanskiy, and Tang}]{sharpness}
\bibinfo{author}{N.~S. Keskar}, \bibinfo{author}{D.~Mudigere}, \bibinfo{author}{J.~Nocedal}, \bibinfo{author}{M.~Smelyanskiy}, \bibinfo{author}{P.~T.~P. Tang},
\newblock \bibinfo{title}{On large-batch training for deep learning: Generalization gap and sharp minima},
\newblock \bibinfo{journal}{arXiv preprint arXiv:1609.04836}  (\bibinfo{year}{2016}).

\end{thebibliography}


\end{document}